%% file: main.tex
\pdfoutput=1

\documentclass[11pt]{article}

\usepackage[preprint]{acl}

\usepackage{times}
\usepackage{latexsym}

\usepackage[T1]{fontenc}

\usepackage[utf8]{inputenc}

\usepackage{microtype}

\usepackage{inconsolata}

\usepackage{booktabs}
\usepackage{graphicx}
\usepackage{subcaption}
\usepackage{amsmath}
\usepackage{amsfonts}
\usepackage{amssymb}
\usepackage[frozencache,cachedir=minted]{minted}
\usepackage[hang,flushmargin]{footmisc}
\usepackage{fontawesome5}
\usepackage{multirow}

\title{LangVAE and LangSpace:\\ Building and Probing for Language Model VAEs}

\author{Danilo S. Carvalho$^{1}$, Yingji Zhang$^{2}$, Harriet Unsworth$^{1}$, Andr\'{e} Freitas$^{1,2,3}$\\
National Biomarker Centre, CRUK-MI, University of Manchester, United Kingdom$^{1}$\\
Department of Computer Science, University of Manchester, United Kingdom$^{2}$ \\
Idiap Research Institute, Switzerland$^{3}$ \\
\faGithub \textit{ }
    https://github.com/neuro-symbolic-ai/\{\href{https://github.com/neuro-symbolic-ai/LangVAE}{LangVAE}, \href{https://github.com/neuro-symbolic-ai/LangSpace}{LangSpace}\} ~|~ 
\faYoutube~~\href{https://youtu.be/DVcrdIX9CfI}{Short Video}
} 

\begin{document}
\maketitle
\begin{abstract}
We present \textit{LangVAE}, a novel framework for modular construction of variational autoencoders (VAEs) 
on top of pre-trained large language models (LLMs). Such language model VAEs can encode the knowledge 
of their pre-trained components into more compact and semantically disentangled representations. 
The representations obtained in this way can be analysed with the LangVAE companion framework: \textit{LangSpace}, 
which implements a collection of probing methods, such as vector traversal and interpolation, disentanglement 
measures, and cluster visualisations. LangVAE and LangSpace offer a flexible, efficient and scalable way of 
building and analysing textual representations, with simple integration for models available on the HuggingFace 
Hub. Additionally, we conducted a set of experiments with different encoder and decoder combinations, as well 
as annotated inputs, revealing a wide range of interactions across architectural families and sizes w.r.t.
generalisation and disentanglement. Our findings demonstrate a promising framework for systematising the experimentation and understanding of textual representations.

\end{abstract}

\input{sections/intro.tex}

\input{sections/llvaes.tex}

\input{sections/langvae.tex}

\input{sections/langspace.tex}

\input{sections/exp_models.tex}

\input{sections/conclusion.tex}


\bibliography{anthology,custom}

\input{sections/appendix.tex}

\end{document}

%% file: sections/intro.tex
\section{Motivation and Purpose}

\begin{figure*}[tb]
  \includegraphics[width=\linewidth]{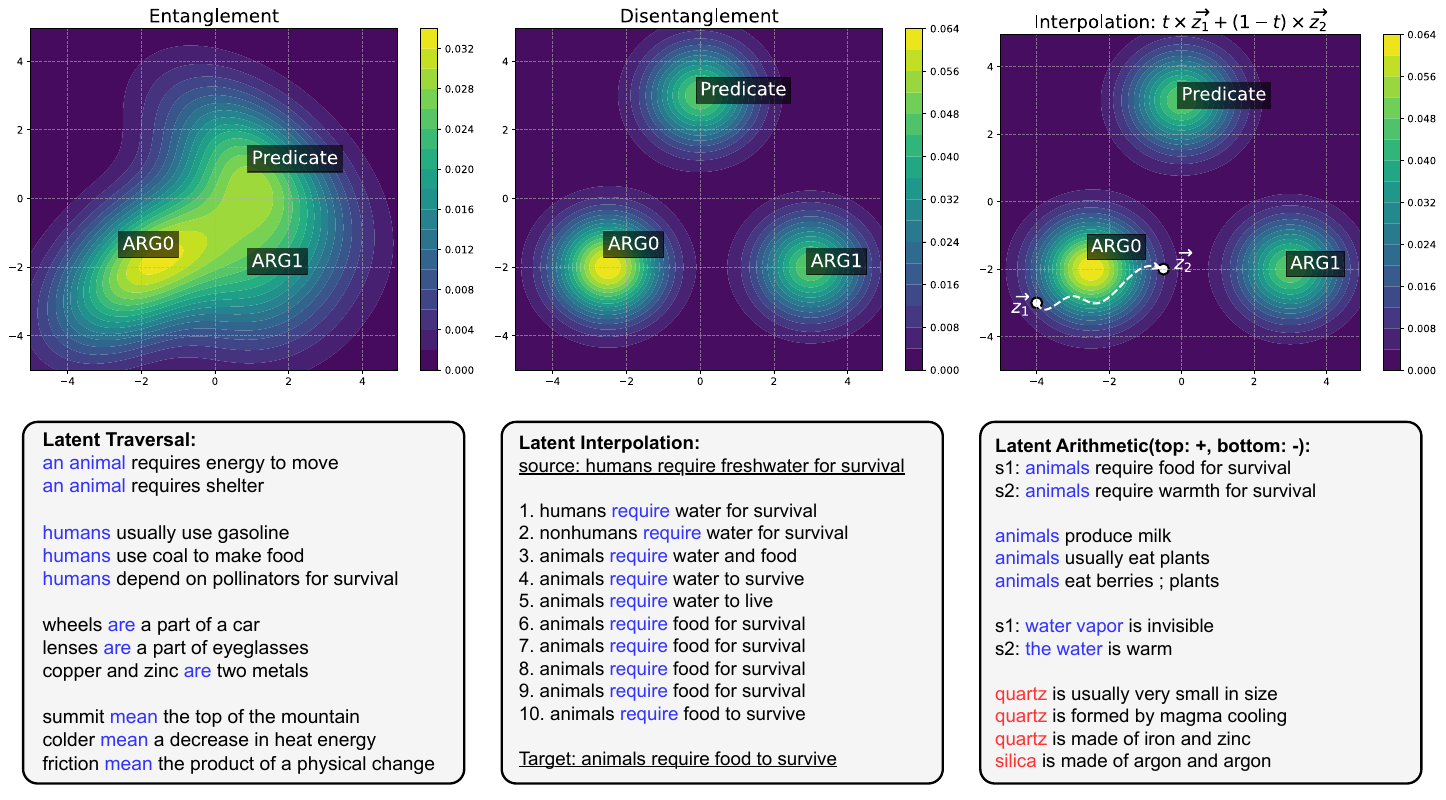}
  \caption{\textit{LangVAE} is a flexible framework designed to support arbitrary combinations of pretrained encoders and decoders for learning latent representations under either a categorical semantic prior or a Gaussian prior. \textit{LangSpace} facilitates comprehensive analysis of the learned latent space through automated evaluation of key properties such as disentanglement and visualization (top) and enables controlled generation by leveraging these latent properties, such as latent traversal, interpolation, and arithmetic operations (bottom).}
  \label{fig:overview}
\end{figure*}

Variational Autoencoders (VAEs) \cite{kingma2013auto} are of considerable importance in machine learning due to their capacity to integrate prior knowledge, quantify uncertainty, enhance generalisation, and deliver interpretability. First, the integration of prior distribution serves as an inductive bias, enabling the model to leverage existing knowledge and providing a principled way to incorporate domain expertise. In the computational linguistics domain, for example, the hierarchical syntax information can be well encoded via hyperbolic prior \cite{davidson2018hyperspherical,cho2023hyperbolic}. Second, their probabilistic formulation allows for explicit uncertainty quantification, providing not only point estimates but also confidence intervals over latent variables and reconstructions, which is significant in the Safety and Trustworthy AI domain, such as hallucinations of LLMs \cite{ji2023survey}. Third, by enforcing a smooth and continuous latent space, VAEs promote better composition and generalisation, as they capture the underlying generative factors of the input distribution \cite{bonnet2024searching}. Fourth, the latent space can compress the knowledge into abstract-level concepts, which is similar to how humans understand the world \cite{barrault2024large}. Concurrently, the rapid pace of development of LLMs has led to substantial gains in a wide variety of NLP tasks, demonstrating remarkable knowledge representation capabilities~\cite{kauf2023event, selby2025had}, but present critical challenges in interpretability and fine-grained control~\cite{kunz2022does, friedman2024interpretability}.

To leverage the strengths of both LMs and VAEs, Language model-based VAEs (LM-VAEs) \cite{bowman2015generating} have been proposed and widely deployed in the controlled text generation domain, such as style transfer tasks: modifying sentences with regard to markers of sentiment, formality, affirmation/negation \cite{bao2019generating,vasilakes-etal-2022-learning,gu-etal-2022-distributional,liu-etal-2023-composable,gu-etal-2023-controllable,liu-etal-2024-multi} and textual, syntactic, semantic representation learning domain \cite{mercatali-freitas-2021-disentangling-generative, de2023learning, zhang2024improving, zhang2024graph}.
However, despite their strategic positioning in delivering more controlled latent representations, there has been limited software infrastructure support to facilitate experimentation with LM-VAEs and in particular, scaling-up to Large Language Model configurations (LLM-VAEs).

In this work we address these issues by presenting a novel framework for modular construction of LM-VAEs on top of pre-trained LMs of different scales, called \textit{LangVAE}, and its companion framework \textit{LangSpace}, dedicated to latent space probing and evaluation. LangVAE introduces a novel approach for latent vector unpooling to autoregressive LMs that sharply reduces the computational and memory requirements, while incorporating compatibility to contemporary LLMs and hardware optimisations.

Finally, we conducted a set of experiments as a case study to demonstrate the frameworks' capabilities and highlight the effects of different combinations of encoder and decoder models, in terms of generalisation and latent space disentanglement, evidencing the impact of facilitating a systematic analysis across different encoder-bottleneck-decoder combinations and parametrisations.


Both frameworks are available as python libraries in the PyPI package repository and on public source code repositories\footnote{\url{https://github.com/neuro-symbolic-ai/LangVAE/}\label{fn:langvae_repo}} \footnote{\url{https://github.com/neuro-symbolic-ai/LangSpace}\label{fn:langspace_repo}}. A demonstration video is available at: \href{https://youtu.be/DVcrdIX9CfI}{youtu.be/DVcrdIX9CfI}.

%% file: sections/llvaes.tex
\section{Language Model VAEs}
A language model VAE (LM-VAE) is a variational autoencoder where both the encoder and decoder components are LMs \cite{bowman2015generating,li-etal-2020-optimus,tu2022adavae,zhang2023llamavae}. It can encode the knowledge of their pre-trained components into compact latent vectors and enables guided language generation from an abstract level using said vectors. The benefits of such models also extend to interpretability (due to their better disentanglement properties), as the VAE architectural bottleneck provides a singular point for probing a model's latent space structure and its syntactic/semantic representation~\cite{li-etal-2020-optimus, mercatali-freitas-2021-disentangling-generative, de2023learning, zhang2024improving, zhang2024graph} and inferential properties \cite{bonnet2024searching}. The creation of continuous latent representation spaces, with better disentanglement and separability of syntactic/semantic properties offers a key mechanism for supporting generative control both at the level of sentences \cite{bao2019generating,felhi2022towards,zhang2024graph} and natural language inferences~\cite{yu2022interaction}. 

\begin{figure}[tb]
  \includegraphics[width=\linewidth]{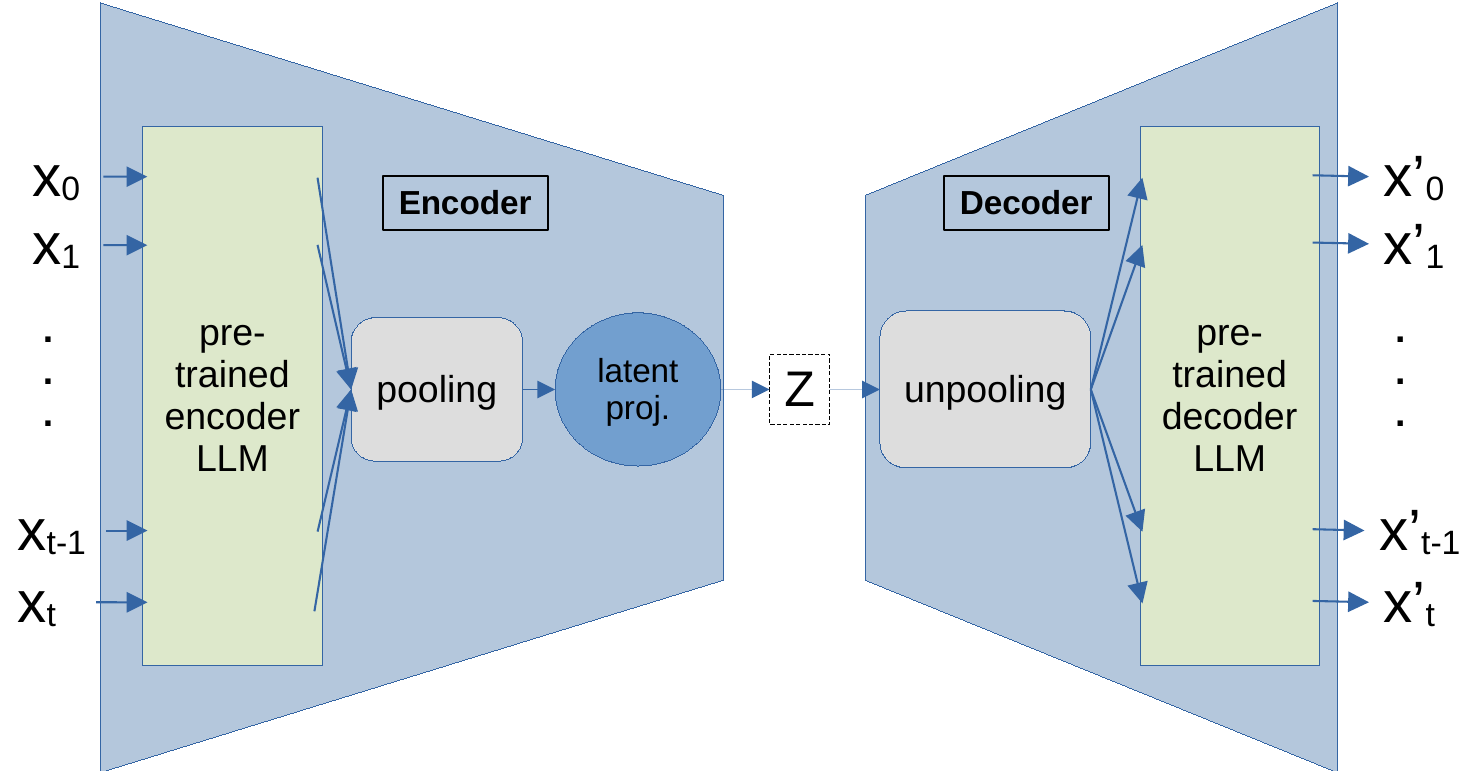}
  \caption{Diagram of fundamental LLVAE architecture.}
  \label{fig:llvae_diag}
\end{figure}

In its most basic conceptualisation, an LM-VAE consists of: (\textbf{a}) an encoder type LLM (e.g., BERT, T5), to provide base representations for each token of the input text; (\textbf{b}) a pooling process to accumulate the input token representations; (\textbf{c}) a projection layer, to convert the base encoding to the regularised VAE latent space; (\textbf{d}) an unpooling process, to derive token representations from a latent vector and feed them to the decoder; and (\textbf{e}) a decoder type LLM (e.g., GPT, Llama) capable of generating tokens from a sequence of input representations. This structure is illustrated in Figure~\ref{fig:llvae_diag}. On the top of this base configuration, syntactic and semantic features can be injected into the latent space, aiming to improve the localisation and control of such features via conditionalisation mechanisms, such as CVAE or clustering losses. Moreover, further architectural interventions can be integrated aiming for additional control, such as the addition of INN layers~\cite{zhang2024learning}, aiming for improving the separability of semantic features.

  \subsection{Optimus}
  The pioneer LLVAE is Optimus~\cite{li-etal-2020-optimus}, which combines a BERT encoder and a GPT-2 decoder to perform sentence encoding, using a mean pooling process, a linear projection layer (MLP), and a unpooling process consisting of two concurrent schemes for latent memory injection to the decoder:
  
  \noindent
  \textit{Memory}: appends a projection of the latent vector directly to each hidden layer of the decoder as a hidden memory vector for the decoder to attend.
  
  \noindent
  \textit{Embedding}: adds a projection of the latent vector to the decoder embedding layer at each decoding step.

  Optimus is trained end-to-end, meaning that the encoder projection layer and the memory and embedding injection layers are jointly trained with the base encoder and decoder models. In this way, the pre-trained models are fine-tuned to "weld" with the projection and injection layers, facilitating convergence. 

  Despite its demonstrated capabilities and potential, Optimus has some limitations, in particular regarding model coupling and scalability. We next discuss our proposed approach for building LM-VAEs and its improvements over the SOTA.


 


%% file: sections/langvae.tex
\section{LangVAE: Building modular LM-VAEs}
\begin{figure}[t]
  \includegraphics[width=\linewidth]{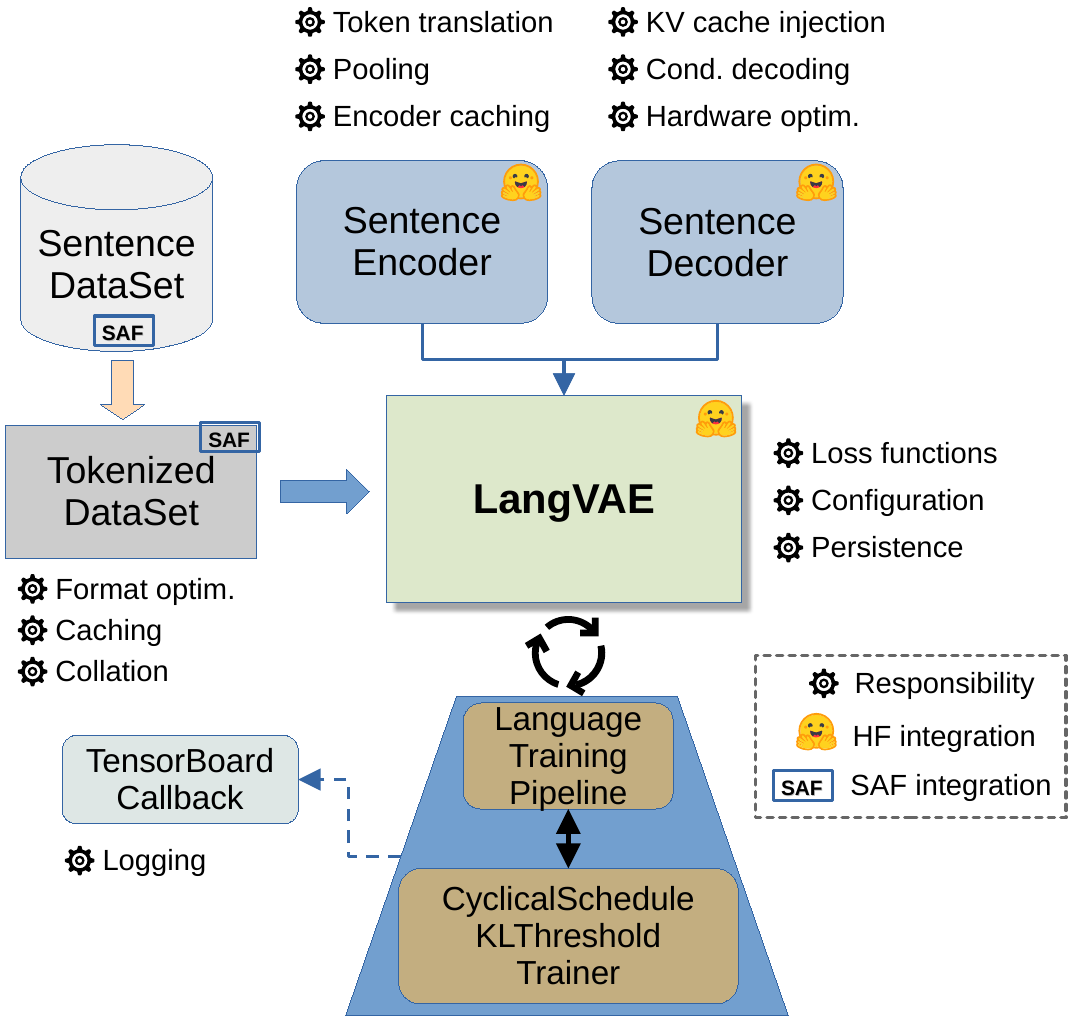}
  \caption{Overview of the LangVAE framework.}
  \label{fig:langvae_diag}
\end{figure}

Aiming to address current LM-VAE limitations and facilitate the development of specialised models and experimentation over next-gen LLMs, we developed \textit{LangVAE}. This is a novel framework specifically targeted at LM-VAE research, focused on the modular development of the architectural components discussed in the previous section (especially projections and unpooling processes), and having a strong integration with the python transformers library\footnote{\url{https://github.com/huggingface/transformers}}. LangVAE is developed and distributed as a python library\footref{fn:langvae_repo} under the GPLv3 License. It is built on top of the pythae library for autoencoders~\cite{chadebec2022pythae}. Figure~\ref{fig:langvae_diag} provides an overview of LangVAE's modules and responsibilities.

  \subsection{Architecture}

  LangVAE implements the fundamental LM-VAE architecture (Figure~\ref{fig:llvae_diag}) in the following ways:

  \paragraph{Pre-trained LLM encoder:} as a loader for an encoder type LLM compatible with the transformers library, via the automodel classes (AutoModel, AutoModelForTextEncoding).
  
  \paragraph{Pooling process:} mean pooling, last hidden state of the base encoder, or the CLS token hidden state, which is automatically selected depending on the pre-trained encoder model configuration.
  
  \paragraph{Latent projection layer:} a linear MLP that adjusts the input encoding size on training time.

  \paragraph{Unpooling process:} a variation of the memory injection scheme from Optimus, called \textit{KV cache injection}, which does not require customisation of the pre-trained decoder code. Instead, it uses the transformers library KV caching mechanism for guiding the decoder (detailed in the next section).

  \paragraph{Pre-trained LLM decoder:} same as the encoder, but relying on the transformers \textit{AutoModelForCausalLM} class for model parametrisation regarding tokenizer configuration and hardware optimisations (e.g., flash attention and multi-GPU distribution).\\

  In addition, LangVAE provides the following functionalities:

  \paragraph{Data conversion:} TokenizedDataSet classes for convenient and efficient tokenisation of text datasets, including the handling of annotations.
  \paragraph{Training pipeline:} supporting cyclical schedule KL annealing to avoid the KL vanishing problem, with beta and KL thresholds.
  \paragraph{Training monitoring:} with tensorboard logging.

  \subsection{KV cache injection}
\begin{figure}[ht!]
    \includegraphics[width=\linewidth]{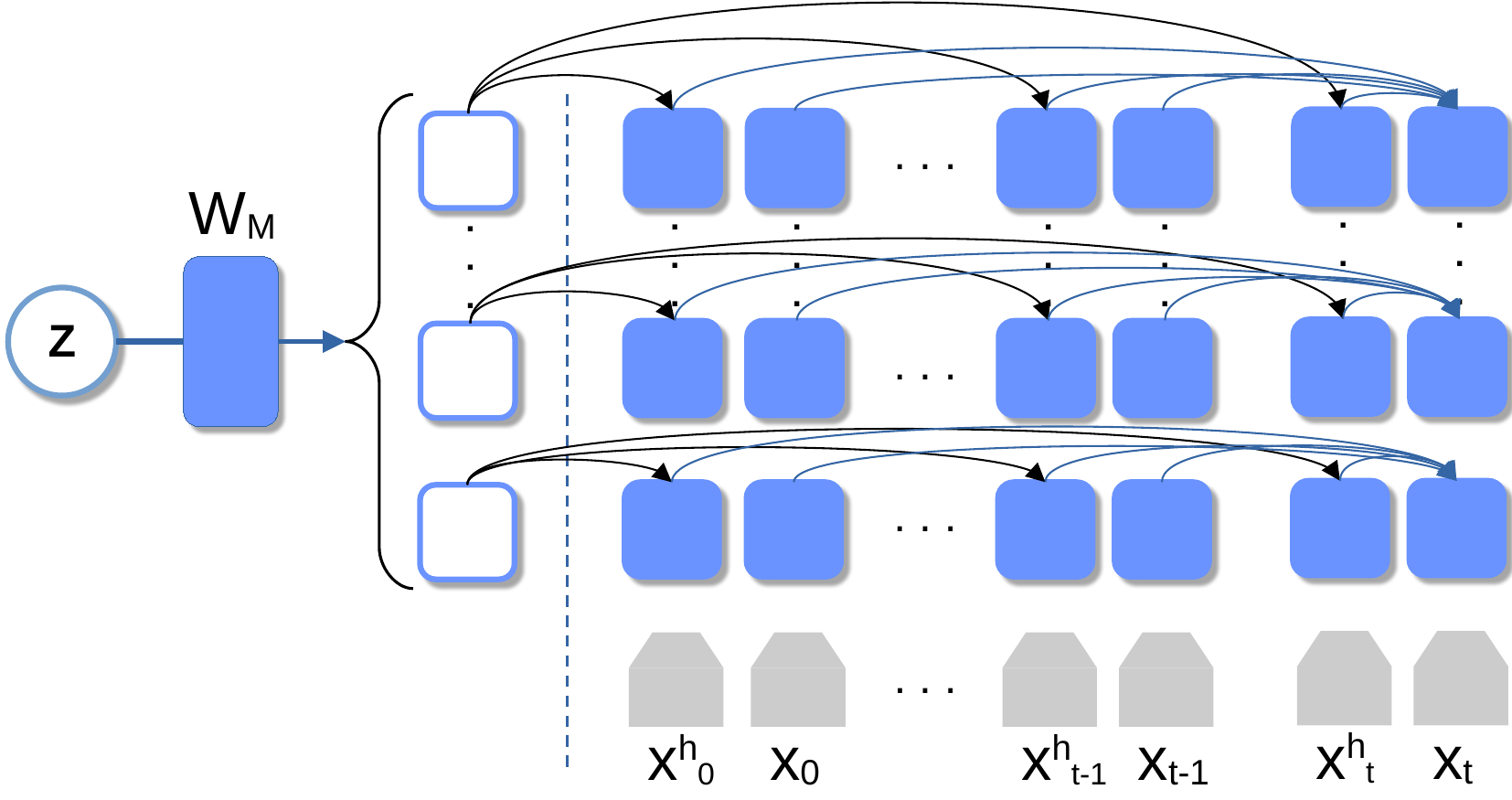}
    \caption{Illustration of the KV cache injection scheme. $W_m z$ projects hidden KV cache entries $X^{h}_{t}$ that are attended by the decoder when predicting the next token. The hidden cache entries are interleaved with the ones produced by the decoder.}
    \label{fig:kv_cache_injection}
  \end{figure}

  One of the central contributions behind LangVAE is the key-value (KV) cache injection scheme, as an alternative to Optimus' memory injection. This new scheme uses the KV caching mechanism of the Causal LM model classes within the transformers library to inject a positional projection of the latent vector. A linear projection of the latent vector $h_{cache} = W_m z$ plays the role of an additional context to guide generation, in the form of hidden KV cache entries $X^{h}_{t}$ interleaved with those produced by the decoder, where $W_m \in \mathbb{R}^{LH \times S}$ is separated into $S \times L$ (sequence length * \# layers) vectors of hidden size $H = K \times V$. Figure~\ref{fig:kv_cache_injection} illustrates this scheme.

  There are two main advantages to this approach. Firstly, it eliminates the need to change the layout of the hidden layers to accommodate the injected memory vector. Therefore, it is compatible with any model that supports the KV caching mechanism. Lastly, in enables training the LM-VAE model with the weights for the base pretrained models frozen, greatly reducing the computational and memory requirements. Additionally, this scheme allows distributed training of the injection layer, as the projection matrices can be co-located with the respective hidden layers in training time, and size of context (number of hidden cache entries) can be adjusted.

  \subsection{Main advantages \& Limitations}
  The main advantages of LangVAE can be summarised as follows:

  \noindent
  • Modular architecture allows flexible development of different LM-VAE configurations. Flexible composition of base models and bottleneck parametrisations, loss functions, etc.

  \noindent
  • Compatible with most state-of-the-art autoregressive models.

  \noindent
  • Has a substantially reduced computational requirements for training, compared to the SOTA LM-VAE (Optimus), with an average parameter reduction of over 95\% measured when using decoder models between 3B to 7B parameters (Section~\ref{sec:exp_setup}).

  \noindent
  • Supports multi-GPU training and inference.\\

  Its main limitations are related to the cache injection mechanism:

  \noindent
  • Slower convergence, as there are far less parameters to adjust.

  \noindent
  • Latent vector sizes tend to be larger, compared to Optimus, to compensate for the overall parameter reduction.\\

  \subsection{Installation and API Examples}
  LangVAE can be installed directly from the PyPI package repository with:  $pip~install~langvae$

  We briefly illustrate the key components of LangVAE's API and how they are instantiated in the supplementary material (Appendix Section~\ref{sec:api_langvae}). A full example of model training can be found in the README file of the code repository\footref{fn:langvae_repo} and on the supporting python notebook\footnote{\url{https://bit.ly/3FMPg5N}}.

%% file: sections/langspace.tex
\section{LangSpace: Simplified probing for LM-VAEs}
 \textit{LangSpace}\footref{fn:langspace_repo}, is a companion framework to LangVAE focused on the evaluation and on the latent space probing for LM-VAEs. It provides an easy-to-use API to perform a variety of analyses on pretrained LM-VAEs models, namely:

\noindent
• Probes: vector arithmetic and interpolation, latent space traversal, disentanglement and cluster visualisation.

\noindent
• Metrics: disentanglement (z-diff, z-min-var, MIG, Disentanglement, Informativeness, Completeness), interpolation (quality, smoothness).\\

  \subsection{Installation and API Examples}
  Like LangVAE, LangSpace can be installed from the PyPI repository with: $pip~install~langspace$

  We briefly illustrate below the use of one of LangSpace probes: latent traversal. A full example with all the available probes can be found within this public notebook\footnote{\url{https://bit.ly/424bjw3}}. 

\begin{minted}{python}
# Seed sentences to traverse
sentences = [
    "animals require food to survive",
    "water vapor is invisible"
]
# Dataset importer
ds = ListImporter()(
    [sent.split() for sent in sentences]
).sentences
# Create dataset tokeniser
seeds = TokenizedDataSet(
    ds, model.decoder.tokenizer, 
    model.decoder.max_len
)
# Create probe and generate report: a
# dataframe with collumns for the 
# original sentence, traversed dimension,
# distance traversed and the generated 
# result, respectively.
trav_report = TraversalProbe(
    model, trav_dataset, 
    sample_size=10, 
    dims=list(range(128))
).report()
\end{minted}

%% file: sections/exp_models.tex
\section{Case study \& Model availability}
To demonstrate LangVAE and LangSpace capabilities and highlight the effects of different combinations of encoder and decoder models, in terms of generalisation and disentanglement of the latent space, we conducted a set of experiments as a case study. The experiments consist of a simple explanation sentence modeling task~\cite{zad2021survey, dalvi-etal-2021-explaining} with posterior evaluation of the induced latent space. Pre-trained checkpoints for all model combinations presented in this study are available in our public HF Hub repository\footnote{\url{https://huggingface.co/neuro-symbolic-ai}}.

  \subsection{Experimental setup} \label{sec:exp_setup}
  For the pre-trained LLMs, we selected three distinct encoder models, in order of parameter size: BERT (base-cased)~\cite{devlin-etal-2019-bert}, Flan-T5 (base)~\cite{chung2024scaling} and Stella (en-1.5B\_v5)~\cite{zhang2025jasperstella}, and four decoder models: GPT-2 (base)~\cite{radford2019language}, Qwen (2.5-3B)~\cite{qwen2.5}, Llama (3.2-3B)~\cite{grattafiori2024llama} and Mistral (7B-v0.3)~\cite{jiang2023mistral7b}. The selection considered the inclusion of different model families and sizes. For each combination, inputs without and with semantic role labeling (SRL) annotations were used (as semantic features), where the SRL annotations were passed as additional variables (one-hot encoded) to the encoder only, going through a separate pooling process (always mean pooling). The latent size (128) and maximum sentence was kept the same for all tests. All models were trained for 50 epochs, with $LR = 0.001$, $target\_kl = 2.0$, $max\_beta = 1.0$, 40 beta annealing cycles, and batch size of 50. Disentanglement measurements were obtained using LangSpace's disentanglement probe for the metrics z-diff~\cite{higgins2016beta}, z-min-var~\cite{kim2018disentangling} and Informativeness~\cite{eastwood2018framework}.

  All experiments were performed on a computer with the following specifications: CPU: AMD EPYC 7413 24-Core, GPU: 2x NVIDIA A100-SXM4-80GB, Memory: 200GB. LangVAE allows caching of the base encoder outputs, which causes the training time to be mostly dominated by the base decoder inference time. The shortest training time was of aprox. 1h (GPT-2) and the longest was about 4.5h (Mistral-7B). Training requirements for larger decoders scale similarly to inference, with a training run of Phi-4 (14B) also taking about 4.5h to complete. The ratios of the LangVAE trained models' size to the base LLMs was: GPT-2 = 0.547, Qwen2.5-3B = 0.024, Llama3.2-3B = 0.076, Mistral-7B = 0.037. Excluding GPT-2, this represents an over 95\% parameter reduction.

  \subsection{Data}
  The same data was used for all tests: a subset of all explanatory sentences from the EntailmentBank dataset~\cite{dalvi-etal-2021-explaining}, which was loaded using the \textit{saf-datasets}\footnote{\url{https://github.com/neuro-symbolic-ai/saf_datasets}} library. The dataset contains 12496 sentences, from which 99\% were used for training and 1\% for validation\footnote{The validation split here is just a means to track the training progress for overfitting, since the dataset is small.}. Evaluation was performed on a random sample of 200 sentences including the validation set and a small portion of the training set.

  SRL annotation was performed using the AllenNLP\footnote{\url{https://github.com/allenai/allennlp-models}} library with a SOTA SRL model~\cite{Shi2019SimpleBM}.

  \subsection{Results}
  The results for the explanation sentence modeling task are presented in Table~\ref{tab:results}. The first observation is that the highest reconstruction performance was achieved by the smallest model combination (for SRL). While not the expected outcome, this can be explained by the constraint imposed on the latent space size, composed with the limited training data, causing the simpler model to better generalise the inputs. 
  
  The encoder complexity has a substantial impact on the generalisation capability of the model: even though bert-base-cased and flan-t5-base have the same encoding size (768), BERT outperforms T5 in most cases, indicating a higher level of information entanglement on T5. Stella, on the other hand, has a much larger encoding size (1536), with a larger dominating effect based on the information loss over the dimensionality reduction.

  The injection of the SRL categories within the model improved reconstruction performance in all combinations except when Mistral is the decoder. This is a surprising result and indicate some particularity of Mistral's internal representations that invite further investigation.
  
  Finally, the SRL categories did not induce consistent improvements on the disentanglement scores, with the exception of Llama3.2, where it led to qualitative improvements, as illustrated in  Figure~\ref{fig:tsne} (Appendix~\ref{sec:qualitative}).

  \begin{table}[t!]
    \centering
    \small
    \addtolength{\tabcolsep}{-0.5em}
    \begin{tabular}{@{}lcccccc@{}}  
    \toprule
        \textbf{Encoder} & \textbf{Decoder} & \textbf{Annot.} & \textbf{Reconstr.} & \multicolumn{3}{c}{\textbf{Disentanglement}}  \\
    &  &  & (BLEU) & z-diff & z-m-var $\downarrow$ & inform. \\
    \midrule
    BERT & gpt-2 & - & 0.76 & 0.46 & 0.68 & 0.36 \\
    BERT & gpt-2 & SRL & \textbf{0.84} & 0.43 & 0.70 & 0.40 \\
    BERT & Qwen & - & 0.44 & 0.58 & 0.69 & 0.46 \\
    BERT & Qwen & SRL & 0.49 & 0.53 & 0.61 & 0.44 \\
    BERT & Llama & - & 0.65 & \textbf{0.62} & 0.71 & 0.38 \\
    BERT & Llama & SRL & 0.80 & 0.59 & 0.65 & 0.43 \\
    BERT & Mistral & - & 0.81 & 0.51 & \textbf{0.59} & 0.43 \\
    BERT & Mistral & SRL & 0.75 & 0.55 & 0.62 & 0.44 \\
    Flan-T5 & gpt-2 & - & 0.11 & 0.50 & 0.62 & 0.35 \\
    Flan-T5 & gpt-2 & SRL & 0.81 & \textbf{0.62} & 0.67 & 0.42 \\
    Flan-T5 & Qwen & - & 0.19 & 0.52 & 0.69 & 0.39 \\
    Flan-T5 & Qwen & SRL & 0.31 & 0.55 & 0.68 & 0.43 \\
    Flan-T5 & Llama & - & 0.74 & 0.52 & 0.68 & \textbf{0.49} \\
    Flan-T5 & Llama & SRL & 0.80 & 0.59 & 0.64 & 0.41 \\
    Flan-T5 & Mistral & - & 0.78 & \textbf{0.62} & 0.61 & 0.39 \\
    Flan-T5 & Mistral & SRL & 0.72 & 0.51 & 0.63 & 0.43 \\
    Stella & gpt-2 & - & 0.18 & 0.50 & 0.68 & 0.34 \\
    Stella & gpt-2 & SRL & 0.61 & 0.52 & 0.65 & 0.40 \\
    Stella & Qwen & - & 0.15 & 0.48 & 0.69 & 0.44 \\
    Stella & Qwen & SRL & 0.27 & 0.54 & 0.66 & 0.43 \\
    Stella & Llama & - & 0.45 & 0.51 & 0.73 & 0.40 \\
    Stella & Llama & SRL & 0.64 & \textbf{0.62} & 0.72 & 0.42 \\
    Stella & Mistral & - & 0.57 & 0.54 & 0.72 & 0.46 \\
    Stella & Mistral & SRL & 0.55 & 0.51 & 0.71 & 0.39 \\
    \bottomrule
    \end{tabular}
    \caption{Results from the explanation sentence modeling experiments. Best values for each column in bold.}
    \label{tab:results}
\end{table}

%% file: sections/conclusion.tex
\section{Conclusion}
In this work we presented \textit{LangVAE}, a modular and efficient library for building language model VAEs (LM-VAEs), and its companion framework \textit{LangSpace}, dedicated to LM-VAE latent space control, probing and evaluation. With the goal of lowering the experimental barriers in this research area, it introduces a novel approach for latent vector unpooling to autoregressive LMs that sharply reduces the computational and memory requirements for training such models, along with a flexible code architecture which is oriented towards modern LLM development.

We demonstrated the capabilities of LangVAE and LangSpace with a set of experiments using different encoder and decoder combinations, as well as annotated inputs, which reveal a wide range of interactions across architectural families and sizes w.r.t. generalisation and disentanglement. Such interactions point to uncovered factors regarding the models' internal representation properties and how they exchange information.

%% file: sections/appendix.tex
\newpage

\appendix

\section{API examples}

\subsection{LangVAE}\label{sec:api_langvae}

\begin{minted}{python}
# Creates a GPT-2 based decoder expecting 
# a latent vector of size 128, that 
# generates a maximum of 32 tokens, 
# distributed on any number of CUDA GPUs. 
decoder = SentenceDecoder(
    "gpt2", latent_size=128, 
    max_len=32, device="cuda", 
    device_map="auto"
)

# Creates a BERT based encoder producing 
# a latent vector of size 128, expecting 
# GPT-2 tokenised inputs. 
encoder = SentenceEncoder(
    "bert-base-cased", latent_size=128, 
    decoder.tokenizer, device="cuda"
)

# Defines a basic VAE model configuration
model_config = VAEConfig(latent_dim=128)

# Initialise LangVAE model
model = LangVAE(
    model_config, encoder, decoder
)

# Alternatively, loads a pretrained 
# checkpoint from the HF Hub.
org = "neuro-symbolic-ai"
name="eb-langvae-flan-t5-base-gpt2-l128"
model = LangVAE.load_from_hf_hub(
    f"{org}/{name}"
)
\end{minted}

\newpage

\section{Qualitative results}\label{sec:qualitative}
\vspace{-6cm}
We present here qualitative results that did not fit in the main text.

\vspace{-2cm}

\begin{figure}[ht!]
    \begin{subfigure}{.5\textwidth}
        \centering
        \includegraphics[width=\linewidth]{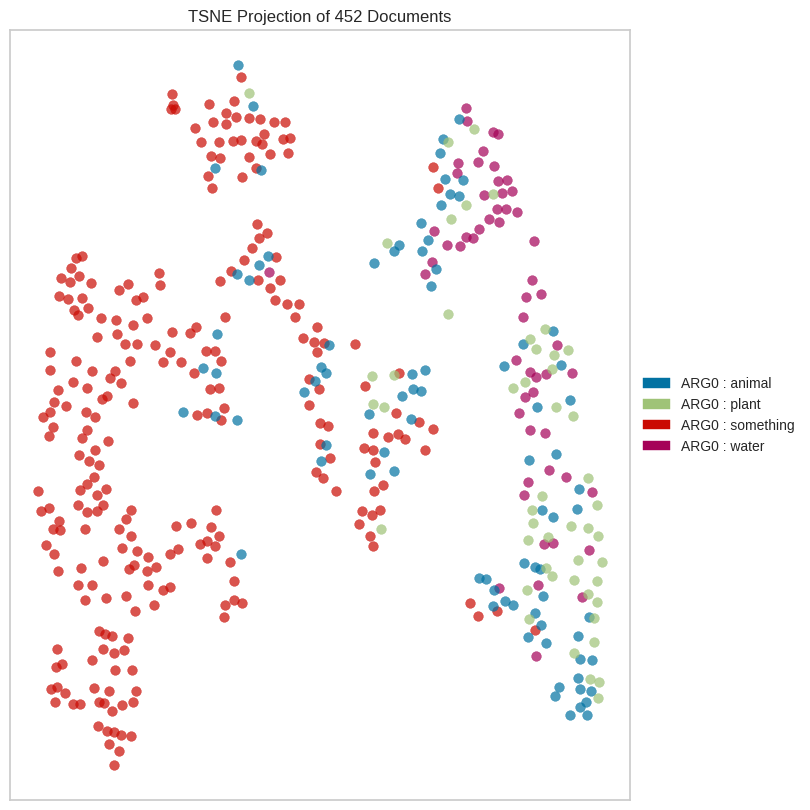}
        \caption{No annotation}
        \label{fig:tsne_a}
    \end{subfigure}
    \begin{subfigure}{.5\textwidth}
        \centering
        \includegraphics[width=\linewidth]{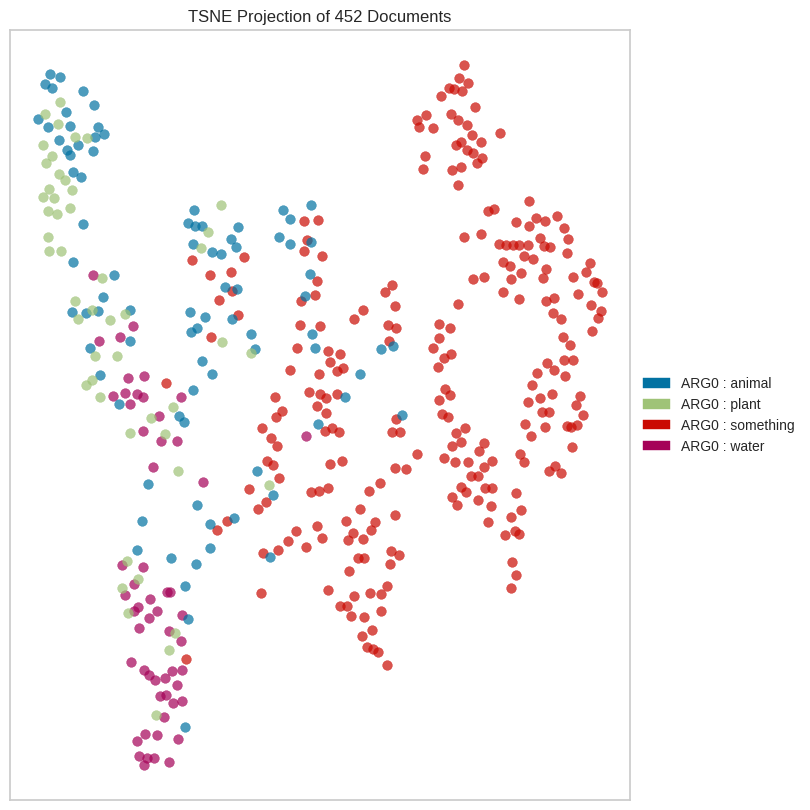}
        \caption{SRL annotation}
        \label{fig:tsne_b}
    \end{subfigure}
    
    \caption{TSNE plots for the [bert-base-cased, Llama-3.2-3B] combination, without (a) and with (b) SRL annotated inputs. We can observe a better separation of the \textit{water} and \textit{animal} subjects on the annotated model.}
    \label{fig:tsne}
\end{figure}

\begin{table}[H]
\small
\addtolength{\tabcolsep}{-0.2em}
\begin{tabular}{llcl}
\toprule
\textbf{Source}                          & \textbf{Target}                          & \textbf{Distance}                & \textbf{Generate}  \\
\midrule
\multirow{11}{*}{the high seas} & \multirow{10}{*}{the continent} & \multirow{11}{*}{0.361} & 1. the high  \\
                                &                                 &                         & 2. the high  \\
                                &                                 &                         & 3. the sea   \\
                                &                                 &                         & 4. the sea   \\
                                &                                 &                         & 5. the sea   \\
                                &                                 &                         & 6. the sea   \\
                                &                                 &                         & 7. the sea   \\
                                &                                 &                         & 8. the land  \\
                                &                                 &                         & 9. the world \\
                                &                                 &                         & 10. the world \\
                                &                                 &                         &           \\
\midrule
                                &                                 &                         &           \\
\multirow{11}{*}{a primary schooler} & \multirow{11}{*}{a college student} & \multirow{11}{*}{0.299} & 1. a primary school  \\
                                &                                 &                         & 2. a primary school  \\
                                &                                 &                         & 3. a junior school   \\
                                &                                 &                         & 4. a junior school   \\
                                &                                 &                         & 5. a high school   \\
                                &                                 &                         & 6. a high school   \\
                                &                                 &                         & 7. a high student   \\
                                &                                 &                         & 8. a college  \\
                                &                                 &                         & 9. a college \\
                                &                                 &                         & 10. a student \\

\end{tabular}

\caption{Interpolation example using a LangVAE model trained on the Wiktionary dataset. Ten points connecting from the source to the target latent vectors are decoded to generate a list of interpolated sentences. We can observe a semantic progression when connecting terms for which there are intermediate senses.}
\end{table}